\documentclass{ieeeaccess}
\usepackage{cite}
\usepackage{amsmath,amssymb,amsfonts}
\usepackage{algorithmic}
\usepackage{graphicx}
\usepackage{textcomp}

\usepackage{times}
\usepackage{soul}
\usepackage{url}
\usepackage[hidelinks]{hyperref}
\usepackage[utf8]{inputenc}
\usepackage[small]{caption}
\usepackage{subfig}
\usepackage{amsthm}
\usepackage{amssymb}
\usepackage{multirow}
\usepackage{booktabs}
\usepackage{algorithm}
\usepackage[switch]{lineno}

\usepackage{bm}
\makeatletter
\AtBeginDocument{\DeclareMathVersion{bold}
\SetSymbolFont{operators}{bold}{T1}{times}{b}{n}
\SetSymbolFont{NewLetters}{bold}{T1}{times}{b}{it}
\SetMathAlphabet{\mathrm}{bold}{T1}{times}{b}{n}
\SetMathAlphabet{\mathit}{bold}{T1}{times}{b}{it}
\SetMathAlphabet{\mathbf}{bold}{T1}{times}{b}{n}
\SetMathAlphabet{\mathtt}{bold}{OT1}{pcr}{b}{n}
\SetSymbolFont{symbols}{bold}{OMS}{cmsy}{b}{n}
\renewcommand\boldmath{\@nomath\boldmath\mathversion{bold}}}
\makeatother

\def\BibTeX{{\rm B\kern-.05em{\sc i\kern-.025em b}\kern-.08em
    T\kern-.1667em\lower.7ex\hbox{E}\kern-.125emX}}

\begin{document}

\title{GINTRIP: Interpretable Temporal Graph Regression using Information bottleneck and Prototype-based method}
\author{{Ali Royat}\authorrefmark{\textdagger}\authorrefmark{,1}, {Seyed Mohamad Moghadas}\authorrefmark{\textdagger}\authorrefmark{,1,2}, Lesley De Cruz\authorrefmark{1,3}, and Adrian Munteanu
\authorrefmark{1,2} \IEEEmembership{Member, IEEE}}

\address[1]{Vrije Universiteit Brussel, Pleinlaan 2, 1050 Brussels, Belgium (e-mail: ali.royat, seyed.mohamad.moghadas,lesley.de.cruz,adrian.munteanu\}@vub.be)}
\address[2]{IMEC, Kapeldreef 75, B-3001 Leuven, Belgium}
\address[3]{Royal Meteorological Institute, Ringlaan 3, 1180 Brussels, Belgium}

\tfootnote{\textdagger~\textbf{These authors contributed equally to this work.}}



\begin{abstract}
Deep neural networks (DNNs) have demonstrated remarkable performance across various domains, but their inherent complexity makes them challenging to interpret. This is especially true for temporal graph regression tasks due to the complex underlying spatio-temporal patterns in the graph. 

While interpretability concerns in Graph Neural Networks (GNNs) mirror those of DNNs, no notable work has addressed the interpretability of temporal GNNs to the best of our knowledge. Innovative methods, such as prototypes, aim to make DNN models more interpretable. However, a combined approach based on prototype-based methods and Information Bottleneck (IB) principles has not yet been developed for temporal GNNs.
Our research introduces a novel approach that uniquely integrates these techniques to enhance the interpretability of temporal graph regression models. The key contributions of our work are threefold: 

We introduce the \underline{G}raph \underline{IN}terpretability in \underline{T}emporal \underline{R}egression task using \underline{I}nformation bottleneck and \underline{P}rototype (GINTRIP) framework, the first combined application of IB and prototype-based methods for interpretable temporal graph tasks.
We derive a novel theoretical bound on mutual information (MI), extending the applicability of IB principles to graph regression tasks.
We incorporate an unsupervised auxiliary classification head, fostering diverse concept representation using multi-task learning, which enhances the model's interpretability.

Our model is evaluated on real-world datasets like traffic and crime, outperforming existing methods in both forecasting accuracy and interpretability-related metrics such as MAE, RMSE, MAPE, and fidelity. The code is available on \href{https://github.com/moghadas76/GINTRIP}{\underline{Github}}.
\end{abstract}


\titlepgskip=-21pt

\maketitle

\section{Introduction}
\textcolor{black}{The rapid expansion of intelligent vehicles, driven by socioeconomic factors and urbanization, underscores the critical necessity for Intelligent Transportation Systems to mitigate the resulting challenges of escalating urban congestion and environmental degradation~\cite{zulfa2024intelligent},~\cite{heryanto2024generic},~\cite{syahbana2023improved},~\cite{effendy_2016},~\cite{5_Adhinata_2021}.}

In the intelligent transportation industry, ensemble tree-based methods such as XGBoost \cite{10.1145/2939672.2939785} are popular for the regression task thanks to their interpretability. Meanwhile, with the success of Graph Neural Networks (GNNs) in a wide range of deep learning tasks, there has
been an increasing demand to explore the decision-making process of these models and provide explanations for their predictions. Existing traffic dashboards such as STRADA \cite{10591664} cannot interpret their predictions.\\ 
Interpretability is one of the most important yet often neglected aspects of deep neural networks (DNNs) that still need more investigation due to their black-box nature. Different tools and methods have been used to unveil this black box. One of them is the Information Bottleneck (IB) \cite{7133169}, a principle for analyzing DNNs using information theoretic concepts. The main advantage of the IB is that a DNN model filters out the noise and extracts informative latent representations that are easier to understand and analyze. This compression facilitates the identification of which aspects of the input data are most influential in the model's predictions, thereby enhancing interpretability\cite{7133169}. On the other hand, techniques like prototype-based methods use learnable blocks that can capture key features that are useful for interpretability \cite{NEURIPS2019_adf7ee2d,schlichtkrull2021interpreting}. For instance, \cite{9476978} attempted to optimize IB bounds via adversarial training for semantic segmentation tasks while the interpretability was neglected. On the other hand, \cite{zhou2022rethinking} tackled the same task with the notion of prototypes, leading to interpretability.
\par In the area of graphs, there are prominent works to integrate IB. Specifically, \cite{luo2020parameterized,ying2019gnnexplainer} are the pioneers who derive tractable bounds to achieve interpretable GNNs. In the terminology of interpretability, one can distinguish the \textit{post-hoc} methods, which help to understand and interpret the predictions of black-box models like deep neural networks \cite{9875989,8954227}, which are often opaque in terms of decision-making processes. Another class of interpretable models is that of the \textit{self-explainable} methods. One such approach is to utilize the IB principle to explore the natural interpretability and generalization capabilities of GNNs \cite{miao2022interpretable}. Furthermore, recent progress in intrinsic approaches \cite{wu2022discovering} has tackled explainability issues and addressed challenges in managing graph out-of-distribution cases by applying invariant learning and causal inference techniques. In the temporal graph domain, IB methods are underexplored. One of the main works in this area is \cite{10.1145/3583780.3614871}, which generates spatial and temporal subgraphs as explainability output. These subgraphs are not linked to prototypes, hindering high-level interpretability.

In this study, we tackle the aforementioned challenges by presenting the GINTRIP framework. Our framework offers temporal graph prediction and interpretability as a self-explainable model. To the best of our knowledge, we are the first to propose this setting. In summary, our contributions are as follows:
\begin{itemize}
    \item We are the first model to perform interpretable temporal graph regression based on the prototype notion.
    \item We provide pseudo-labels to facilitate interpretability.
    \item Our framework derives tractable \textcolor{black}{mutual information (MI)} bounds for the temporal graph regression problems.
    \item Our model is extensively evaluated on real-world traffic and crime datasets, for which it outperforms the state-of-the-art.
\end{itemize}

\begin{figure*}[htbp]
\centerline{\includegraphics[width=\linewidth,keepaspectratio]{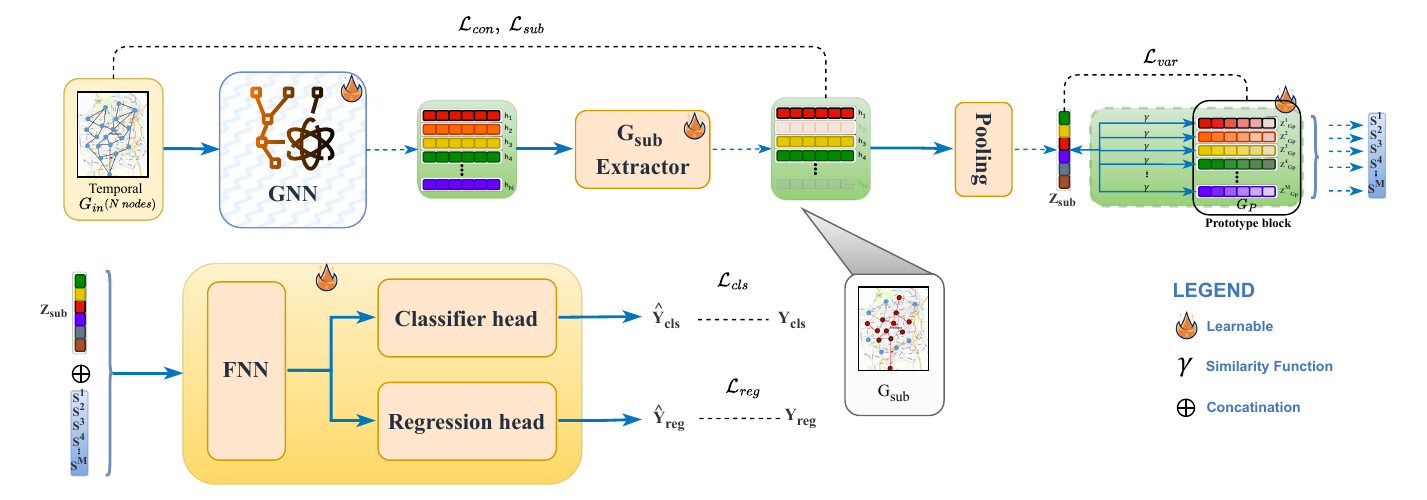}}
\caption{Proposed method architecture. It includes a $GNN$ encoder, subgraph extractor, learnable prototypes, regression, and classification head components. Overall, the temporal graph passes to the $GNN$ model, and the latent representation  \{$h_{1},h_{2},...,h_{N}$\} is processed by $G_{sub}$ $ Extractor$ to extract a subgraph $G_{sub}$. A pooling block is used to capture global-level features. A similarity function $\gamma$ computes similarities between 
learnable prototypes ${G}_{p} = \left\{ \mathbf{z}_{G_{p}}^1, \mathbf{z}_{G_{p}}^2, \cdots, \mathbf{z}_{G_{p}}^M \right\}$
and subgraph representation $Z_{sub}$. The concatenated similarities \{${S^1,S^2,...,S^M}$\} with $Z_{sub}$ are used for the proposed pseudo-label classification and the forecasting task.}
\label{fig}
\end{figure*}

\section{Methodology}

\subsection{Problem Statement} \label{problem}

We focus on temporal graph regression, a technique for predicting time-dependent node values in graph-structured data. We apply this to the datasets, which anticipate traffic and crime conditions based on historical data.
In our graph model $G_{in} = (V_{G_{in}}, E_{G_{in}}, A_{G_{in}})$. Here, $V_{G_{in}} = \left\{V_1, V_2, ..., V_N\right\}$ is the set of nodes, $E_{G_{in}}$ the set of edges, and $A_{G_{in}} \in \mathbb{R}^{N \times N}$ the adjacency matrix. At each timestamp $t$, the graph $G_{in}$ includes a dynamic feature matrix $X_t \in \mathbb{R}^{N \times D}$. In this study, the graph structure is static, and the node's features are attributes related to each dataset, e.g., traffic speed in the traffic dataset. Given a forecast horizon of length $T^{\prime}$ and an observed series history of length $T$, $\left[ y_1, . . . , y_T \right] \in \mathbb{R}^T$, the goal is to predict the vector of future values. For simplicity, we consider a sliding window of length $W \leq T$ ending with the most recent observed value $y_T$ as the model input, denoted as $\boldsymbol{x} \in \mathbb{R}^W = \left[ y_{T-W+1}, y_{T-W+2},..., y_{T} \right] $, where $\boldsymbol{Y}$ is the prediction series. The prototype is defined as a set ${G}_{p} = \left\{ \mathbf{z}_{G_{p}}^1, \mathbf{z}_{G_{p}}^2, \cdots, \mathbf{z}_{G_{p}}^M \right\}$, where $M$ is the total number of prototypes, and each prototype $\mathbf{z}_{G_{p}}^m$ is a learnable parameter vector that acts as the latent representation of the prototypical part (i.e., ${G}_p$) of graph ${G_{in}}$. We allocate $J$ prototypes for each class, i.e., $M=K\times J$, where $K$ denotes the number of classes corresponding to our unsupervised pseudo-classes in the traffic dataset. Indeed, for each node $V_i$, a threshold $T_i = Q_{0.1}(X_{:,i,:})$ classifies each node into one of our predefined pseudo-classes(congested/non-congested) in the traffic dataset, where $Q(.)$ is quantile function. For the crime dataset, we can use available labels.
\\

\subsection{Graph information bottleneck}

The MI between two random variables of $X, Y$ is defined as follows:

\begin{equation}
\small
I(X;Y) = \int_X \int_Y p(x,y) \log \frac{p(x,y)}{p(x)p(y)} \,dx\,dy
\end{equation}\\

The IB method creates a compressed representation $Z$ from input $X$, preserving essential components for predicting output $Y$ while discarding redundant information. The IB objective is:

\begin{equation}
\small
\min_Z -I(Y;Z) + \beta I(X;Z)
\end{equation}\\

Where $\beta$ is the Lagrange multiplier that controls the trade-off between two terms. The first part is responsible for the compression concept, and the second concerns output prediction.

The IB concept has recently been found to be applicable in graph learning, resulting in the development of the IB-Graph. This approach aims to create a condensed representation of a given graph $G_{in}$ while preserving its essential characteristics. The method draws inspiration from the Graph Information Bottleneck (GIB) principle, which focuses on extracting a compact yet interpretable subgraph $G_{sub}$ from the original graph $G$. This extraction process is guided by an optimization objective designed to balance information retention and compression:

\begin{equation}
\small
\min_{G_{sub}} -I(Y;G_{sub}) + \beta I(G_{in};G_{sub})
\label{eq:GIB}
\end{equation}
\subsection{Prototype Injection}
usepackage[normalem]{ulem}
Prototype-based methods are machine learning techniques that represent data by identifying typical examples, or prototypes, which encapsulate the most relevant features of a dataset. These methods facilitate interpretability by allowing models to make predictions based on the similarity of new inputs to these representative prototypes. In our approach, we integrate the \textcolor{black}{GIB} with prototype-based techniques to extract meaningful subgraphs from input graphs. Specifically, we develop a learnable prototype layer designed to identify predictive temporal patterns. This is achieved by computing the similarity between the extracted subgraphs and a prototype matrix, which serves as a predictive feature during the forecasting process. Using $G_{p}$, we can inject $G_{p}$ learnable variables to the first term of (\ref{eq:GIB}) i.e. \begin{small}$I(Y;G_{sub})$\end{small}. Based on the chain rule of MI, we can write: 

\begin{equation}
\small 
I(Y;G_{sub})=I(Y;G_{sub},G_{p}) - I(Y;G_{p}|G_{sub})
\end{equation}\\
and then we can rewrite (\ref{eq:GIB}): 

\begin{equation}
\small 
\min_{G_{sub}} -I(Y;G_{sub},G_{p}) + I(Y;G_{p}|G_{sub}) + \beta I(G_{in};G_{sub})
\end{equation}\\
Due to some difficulty in computing the second term, we again use the chain rule of MI. We can decompose the second term, and we have a more extended expression:

\begin{equation}
\small
\begin{aligned}
\min_{G_{sub}} &- I(Y;G_{sub},G_{p}) + I(G_{p};Y,G_{sub}) - I(G_{p};G_{sub})\\ 
&+ \beta I(G_{in};G_{sub})
\end{aligned}
\label{eq:PGIB_d1}
\end{equation}\\
Observing the first two terms in (\ref{eq:PGIB_d1}), we recognize that they are opposite, meaning the first MI tends to increase, and the second one tends to decrease. However, after enough training time, these two terms compensate each other, where an increase in \begin{small}$I(Y;G_{sub},G_{p})$\end{small} leads to an increase in the \begin{small}$I(G_{p};Y,G_{sub})$\end{small} that is different behavior than (\ref{eq:PGIB_d1}). We therefore exclude the second MI term from our optimization objective to prevent this misalignment:

\begin{equation}
\small
\min_{G_{sub}} - I(Y;G_{sub},G_{p}) - I(G_{p};G_{sub})+\beta I(G_{in};G_{sub})
\label{eq:PGIB_pref}
\end{equation}\\
For an approximate derivation to support the exclusion of this term in (\ref{eq:PGIB_d1}), at first, we can write:

\begin{equation}
\small
I(G_{p};Y,G_{sub}) = I(Y;G_{sub},G_{p}) + I(G_{p};G_{sub})- I(Y;G_{sub})
\end{equation}\\
Based on (\ref{eq:PGIB_d1}), after sufficient training time, \begin{small}$I(G_{p};G_{sub})$\end{small} will increase, which means that $G_{p}$ and $G_{sub}$ will have more anafter sufficient trainind enough correlation and dependency. Furthermore, from figure \ref{fig}, we can see that the relationship complexity of $G_{p}$ and $G_{sub}$ is significantly lower than that of $G_{sub}$ and \begin{small}$Y=(Y_{cls}, Y_{reg})$\end{small}. So for an asymptotic analysis, we can suppose that $G_{p} \approx G_{sub}$, which gives \begin{small}$I(G_{p};G_{sub}) \approx H(G_{sub})$\end{small} which \begin{small}$H(G_{p})$\end{small} is the entropy of $G_{p}$ and then we will have: \begin{small}$I(G_{p};G_{sub}) \gtrsim I(Y;G_{sub})$\end{small}. So, it follows immediately that \begin{small}$I(G_{p};Y,G_{sub}) \gtrsim I(Y;G_{sub},G_{p}) $\end{small} which means an increase in \begin{small}$I(Y;G_{sub},G_{p})$\end{small} leads to an increase in \begin{small}$I(G_{p};Y,G_{sub})$\end{small}. This is an undesirable and conflicting behavior for (\ref{eq:PGIB_d1}) which intend to decrease; \begin{small}$I(G_{p};Y,G_{sub})$\end{small}. So, we exclude \begin{small}$I(G_{p};Y,G_{sub})$\end{small} from our optimization objective to prevent this misalignment.\\

\subsection{Subgraph Extraction}\label{subgraph}

Our model adopts Spatio-Temporal Graph Neural Networks (STGNN) to process the input temporal graph. More specifically, we use MTGNN\cite{wu_connecting_2020} as our GNN backbone. Following \cite{seo_interpretable_2023}, we pass the nodes' extracted features, referred to as $h_{i}$ in figure \ref{fig}, to the sequence of an MLP layer and a sigmoid activation function. Their output assigns the probability $p_i$ of the node inclusion in the selected subgraph, the output of the $G_{sub}$-Extractor block in figure \ref{fig}. The final representation of each selected subgraph node would be:

\begin{equation}
\small
\small z_i = \lambda_ih_i + (1-\lambda_i)\epsilon
\label{final_representation}
\end{equation}\\
, where \begin{small}$\lambda_i \sim Bernoulli(p_i)$\end{small} and \begin{small}$\epsilon \sim \mathcal{N}(\mu_{h_i}, \sigma_{h_i}^2)$\end{small}. The learned probability $p_{i}$ facilitates selective information retention in $G_{sub}$. 
This approach preserves interpretability within the subgraph while also potentially aiding the learning of prototypes introduced in the previous step. To minimize the MI between the input graph and the selected subgraph, \cite{9880086} showed that it is equivalent to minimizing the following upper bound for $I(G_{in}, G_{sub})$:

\begin{equation}
\small
\begin{aligned}
& I(G_{in}, G_{sub}) \le \mathbb{E}_{G_{in}} \left[ -\frac{1}{2} \log S + \frac{1}{2|V_{G_{in}}|} S + \frac{1}{2|V_{G_{in}}|} M^2 \right] \triangleq \mathcal{L}_{sub}
\label{I(g;gsub)}
\end{aligned}
\end{equation}\\
where $S = \sum_{i=1}^{|V_{G_{in}}|}(1-\lambda_i)^2$ and $M = \frac{ \sum_{i=1}^{|V_{G_{in}}|} \lambda_i(h_i-\mu_{h_i}) }{\sigma_{h_i}}$. After noise adding, we compute the embedding $Z_{sub}$ by adopting mean pooling on the subgraph \cite{bianchi2023expressivepowerpoolinggraph}.

To enforce learning connected subgraphs \cite{seo_interpretable_2023}, we employ the following loss function, which is called \textit{connectivity loss}:

\begin{equation}
\small
\mathcal{L}_{con} \triangleq \left\| \widehat{P^T A P} - I_2 \right\|_F
\label{eq:conectivity}
\end{equation}\\
$P\in \mathcal{R}^{\left|V_G\right| \times 2}$ and $A$ are the node probability assignment and adjacency matrix at the batch level, respectively. $I_2$ is the 2x2 identity matrix, $\|...\|_F$ is the Frobenius norm operation and $\widehat{(...)}$ is the row normalization. More precisely, each $P$ matrix row is related to the probability of its corresponding node, meaning that for each row we have ($p_{i}, 1 - p_{i}$). This preserves locality, which is a crucial aspect of the extracted subgraph \cite{NEURIPS2019_d80b7040}. 




The first MI term in optimization (\ref{eq:PGIB_pref}), contains the variables' group \begin{small}$Y=(Y_{cls}, Y_{reg})$\end{small} which corresponds to our network's two output heads: classification and regression. Using chain rule, we can derive a desired lower bound: \begin{small}$I(Y;G_{sub},G_{p}) \geq \frac{1}{2}I(Y_{cls};G_{sub},G_{p}) + \frac{1}{2}I(Y_{reg};G_{sub},G_{p})$\end{small}. Then, we can rewrite the optimization (\ref{eq:PGIB_pref}) as:

\begin{equation}
\small
\begin{aligned}
&\min_{G_{sub}} - \frac{1}{2}I(Y_{cls};G_{sub},G_{p}) - \frac{1}{2}I(Y_{reg};G_{sub},G_{p}) - I(G_{p};G_{sub}) \\
&+ \beta I(G_{in};G_{sub})
\end{aligned}
\label{eq:PGIB_final}
\end{equation}
\subsection{Theoretical Bounds}\label{theory_bnd}
To optimize our objective, we must consider the inherent characteristics of each MI term. By applying appropriate lower and upper bounds to these terms, we define a constrained optimization problem that forms our final loss function \ref{final_loss}. Below, we derive lower bounds for the first two terms of the optimization objective \ref{eq:PGIB_final}.

\begin{equation}
\small
\begin{aligned}
&I(Y_{cls};G_{sub}, G_p) = \\
&\mathbb{E}_{Y_{cls},G_{sub},G_p}[\log p(Y_{cls}|G_{sub},G_p)] - \mathbb{E}_{Y_{cls}}[\log p(Y_{cls})] \\
&\geq \mathbb{E}_{Y_{cls},G_{sub},G_p}[\log p(Y_{cls}|\gamma(G_{sub},G_p))] - \mathbb{E}_{Y_{cls}}[\log p(Y_{cls})] \\
&\geq\mathbb{E}_{Y_{cls},G_{sub},G_p}[\log q_\phi(Y_{cls}|\gamma(G_{sub},G_p))] \triangleq -\mathcal{L}_{cls}
\end{aligned}
\label{eq:UPB_CLS}
\end{equation}\\
Where $\gamma$ is a similarity function, such as dot product similarity, between \begin{small}$G_{sub},G_p$\end{small}. The first inequality was derived from \textcolor{black}{Data Process Inequality (DPI)}\cite{ash2012information} and the second one, caused by \begin{small}$q_\phi(Y_{cls}|\gamma(G_{sub},G_p))$\end{small}, which is the variational approximation to the true posterior \begin{small}$p(Y_{cls}|\gamma(G_{sub},G_p))$ \end{small}. Equation (\ref{eq:UPB_CLS}) illustrates that maximizing the MI \begin{small}$I(Y_{cls};G_{sub}, G_p)$\end{small} is equivalent to minimizing the classification loss, $\mathcal{L}_{cls}$. In practice, the cross-entropy loss is typically employed for categorical $Y_{cls}$.

The previous derivation can be extended to our regression head, requiring additional mathematical (using Pinsker's inequality) steps. This extension results in a lower bound expressed in terms of MSE:

\begin{equation}
\small
\begin{aligned}
I(Y_{reg};G_{sub}, G_p) \geq -\lambda MSE(\hat{Y}_{reg}, Y_{reg}) \triangleq -\lambda \mathcal{L}_{reg} \\
\end{aligned}
\label{eq:UPB_reg}
\end{equation}\\
Where $\lambda$ is a constant coefficient and also $\hat{Y}_{reg}$ and $Y_{reg}$ are the regression head prediction and the regression actual label (ground-truth), respectively (refer to figure \ref{fig}).

Following a similar approach to that used in equation (\ref{eq:UPB_reg}), we can express:

\begin{equation}
\small
\begin{aligned}
I(G_{sub}; G_p) \geq  -\lambda' MSE(q_\theta(Z_{sub}), Z_{G_p}) \triangleq -\lambda' \mathcal{L}_{var} \\
\end{aligned}
\label{eq:UPB_var}
\end{equation}\\

Here, $\lambda'$ is a constant coefficient. $Z_{sub}$ denotes the output of the pooling function applied to $G_{sub}$, while $Z_{G_p}$ represents the concatenation of all prototype vectors $\left\{ \mathbf{z}_{{G}_{p}}^1, \mathbf{z}_{{G}_{p}}^2, \cdots, \mathbf{z}_{{G}_{p}}^M \right\}$. For $q_\theta$, we employ a single-layer perceptron to estimate $Z_{G_p}$ from $Z_{sub}$.\\

The total loss function is a convex combination of individual losses derived from the aforementioned bounds: 

\begin{equation}
\mathcal{L}=\lambda_1\mathcal{L}_{reg} + \lambda_2\mathcal{L}_{sub} + \lambda_3\mathcal{L}_{var} + \lambda_4\mathcal{L}_{con} + \lambda_5\mathcal{L}_{cls}    
\label{final_loss}
\end{equation}\\
To address the instability of this function during training, we employed the multi-loss variation coefficient adjustment approach \cite{groenendijk2020multilossweightingcoefficientvariations}.
\label{refcoeffadjusment}
\textcolor{black}{Based on this method, each coefficient $\lambda_s$ is computed as
$\frac{1}{z}\frac{\sigma_{\mathcal{L}_s}}{\mu_{\mathcal{L}_s}}$, where
$\mu_{\mathcal{L}_s}$ and $\sigma_{\mathcal{L}_s}$ denote the mean and
standard deviation of the loss term $\mathcal{L}_s$, respectively. The
normalizing constant is defined as
$z=\sum_{s=1}^{5}\frac{\sigma_{\mathcal{L}_s}}{\mu_{\mathcal{L}_s}}$,
which guarantees $\sum_{s=1}^{5}\lambda_s=1$. This constraint is important
for decoupling the loss weighting from the learning rate.}

\section{Experiments}
We evaluate the performance of GINTRIP in terms of forecasting metrics and explainability on three distinct real-world traffic and crime datasets. In terms of explainability metrics, the fidelity is defined as \cite{seo_interpretable_2023}:
\begin{equation}
\small \text{Fidelity}^{+/-} = \frac{1}{\lfloor T/W \rfloor} \sum_{i=1}^{\lfloor T/W \rfloor} \left( |f(G_{i}) - f(G_i^{+/-})| \right)
\end{equation}

$f$ represents the trained predictive spatio-temporal function. $G_i^{-}$ and $G_i^{+}$ refer to the selected subgraph structure and the complementary structure of the selected subgraph in G respectively, for $i^{th}$ $G_{in}$. It directly measures the quality of the learned prototypes on the targeted task. Also, in \textcolor{black}{F}igure \ref{fig_vis}\textcolor{black}{,} density, \textcolor{black}{$k$}, denotes the ratio of important nodes (the number of nodes in subgraph structures) to total nodes in the original graph. Node importance is determined based on the assignment probability explained in Section \ref{subgraph}.

\subsection{Quantification analysis}
We conducted our experiments on real-world traffic datasets. Specifically, PeMS04, PeMS07, and PeMS08 \cite{10.1145/3447548.3467430} were chosen as graph-based traffic datasets. 

To evaluate the proposed method, we compare GINTRIP with two groups of methods:
\begin{enumerate}
    \item STGNN Traffic Methods:
        \begin{itemize}
            \item \textbf{STSGCN}\cite{Zheng_Fan_Wang_Qi_2020}, \textbf{DSTAGNN}\cite{shang2021discrete}: employs gated spatial-temporal graph convolution to proceed the traffic as a spatio-temporal tensor by different aggregation techniques.
        \end{itemize}
    \item Explainable STGNNs:
    \textbf{STExplainer}\cite{10.1145/3583780.3614871}: decomposes the explainability into the spatial and the temporal domain. 
\end{enumerate}

The traffic forecasting metrics are presented in Table \ref{reg_table_pems4}. The results indicate that \textit{self-explainablity} promotes the predictivity capacity. In contrast to \cite{10.1145/3583780.3614871}, our extracted subgraphs are time-aware, which eases the interpretation process. Indeed, it dispenses with the post-processing of temporal and spatial subgraphs proposed in \cite{10.1145/3583780.3614871}. 

\begin{table*}[htbp]
\small  
\setlength{\tabcolsep}{4pt}  
\caption{Traffic predictions benchmark, B.B. is the abbreviation for block box models which are not interpretable}
\begin{center}
\begin{tabular}{@{}l@{\hspace{3mm}}lccccccccc@{}}
\toprule
 & \multirow{2}{*}{Baselines} & \multicolumn{3}{c}{PeMS04} & \multicolumn{3}{c}{PeMS07} & \multicolumn{3}{c}{PeMS08} \\ \cmidrule{3-11}
& & MAE$\downarrow$ & RMSE$\downarrow$ & MAPE(\%)$\downarrow$ & MAE$\downarrow$ & RMSE$\downarrow$ & MAPE(\%)$\downarrow$ & MAE$\downarrow$ & RMSE$\downarrow$ & MAPE(\%)$\downarrow$ \\
\midrule
\multirow{2}{*}{\rotatebox[origin=c]{90}{B.B.}} 
& STGNCDE \cite{Zheng_Fan_Wang_Qi_2020} & 19.21 & 31.09 & 12.76 & 20.53 & 33.84 & 8.80 & 15.45 & 24.81 & 9.92 \\ \cline{2-11}
& DSTAGNN \cite{shang2021discrete} & 19.30 & 31.46 & 12.70 & 21.42 & 34.51 & 9.01 & 15.67 & 24.77 & 9.94 \\ \cline{2-11}
\midrule
\multirow{2}{*}{\begin{tabular}[l]{@{}l@{}}Interpretable\end{tabular}} 
& STExplainer \cite{10.1145/3583780.3614871}  & \underline{19.09} & \underline{30.64} & \textbf{12.21} & \textbf{20.00} & \textbf{33.45} & \textbf{8.51} & \textbf{14.70} & \textbf{23.91} & \textbf{9.80}  \\ \cline{2-11}
& GINTRIP (Ours)  & \textbf{18.62} & \textbf{29.02} & \underline{12.75} & \underline{20.12} & \underline{33.53} & \underline{8.62} & \underline{14.88} & \underline{24.05} & \underline{9.89} \\
\bottomrule
\end{tabular}
\label{reg_table_pems4}
\end{center}
\end{table*}

To validate GINTRIP’s generalizability, we conducted comprehensive experiments across multiple crime datasets, NYC and Chicago\cite{Xia_2021}. The results are presented in Table\ref{tab_crime}. Consistently, GINTRIP outperforms specialized crime prediction methods like DMSTGCN~\cite{10.1145/3447548.3467275} attributed to its interpretability. Based on the superior or same performance, GINTRIP holds up the statement that self-explainability causes predictability and robustness.


\begin{table}
\small
\setlength{\tabcolsep}{4pt}  
\caption{Crime prediction results on NYC and Chicago datasets}
\begin{center}
\begin{tabular}{|l|cc|cc|}
\hline
\multirow{2}{*}{Model} & \multicolumn{2}{c|}{NYC Crime} & \multicolumn{2}{c|}{CHI Crime} \\
\cline{2-5}
& MAE & MAPE & MAE & MAPE \\
\hline
ST-SHN \cite{Xia_2021} & 0.9280 & 0.5373 & 1.0689 & 0.5116 \\ \hline
DMSTGCN \cite{10.1145/3447548.3467275} & 0.9293 & 0.5485 & 1.0736 & 0.5175 \\ \hline
STExplainer \cite{10.1145/3583780.3614871} & \underline{0.9095} & \underline{0.5154} & \underline{1.0701} & \underline{0.5143} \\ \hline
GINTRIP (ours) & \textbf{0.8912} & \textbf{0.5016} & \textbf{1.055} & \textbf{0.5117} \\
\hline
\end{tabular}
\label{tab_crime}
\end{center}
\end{table}

\begin{figure}[htbp]
\centerline{\includegraphics[width=0.8\linewidth,keepaspectratio]{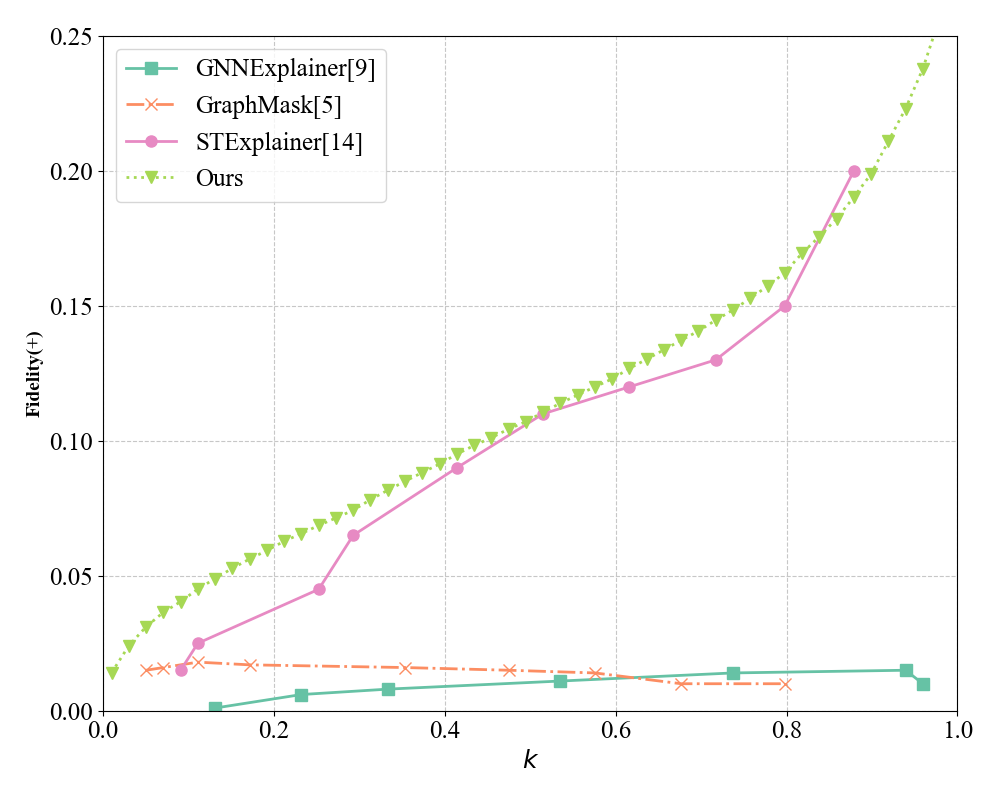}}
\vspace{-0.35cm}
\caption{Positive fidelity over k. High fidelity is an indicator of interpretability.}
\label{fig_fid}
\end{figure}

To verify our interpretability, we compare the fidelity and $K$ with STExplainer\cite{10.1145/3583780.3614871}, which is presented in figure \ref{fig_fid}. Our $Fidelity^+$ score is superior to the previous state-of-the-art. In the figure \ref{fig_fid}, the x-axis is the density parameter. So, mostly, with the different degrees of the subgraph density, the $Fidelity^+$ of our model is better than STExplainer\cite{10.1145/3583780.3614871}, implying superior interpretability.

\vspace{-0.15cm}
\subsection{Visualizations}
\begin{figure}[htbp]
\centerline{\includegraphics[width=\linewidth,keepaspectratio]{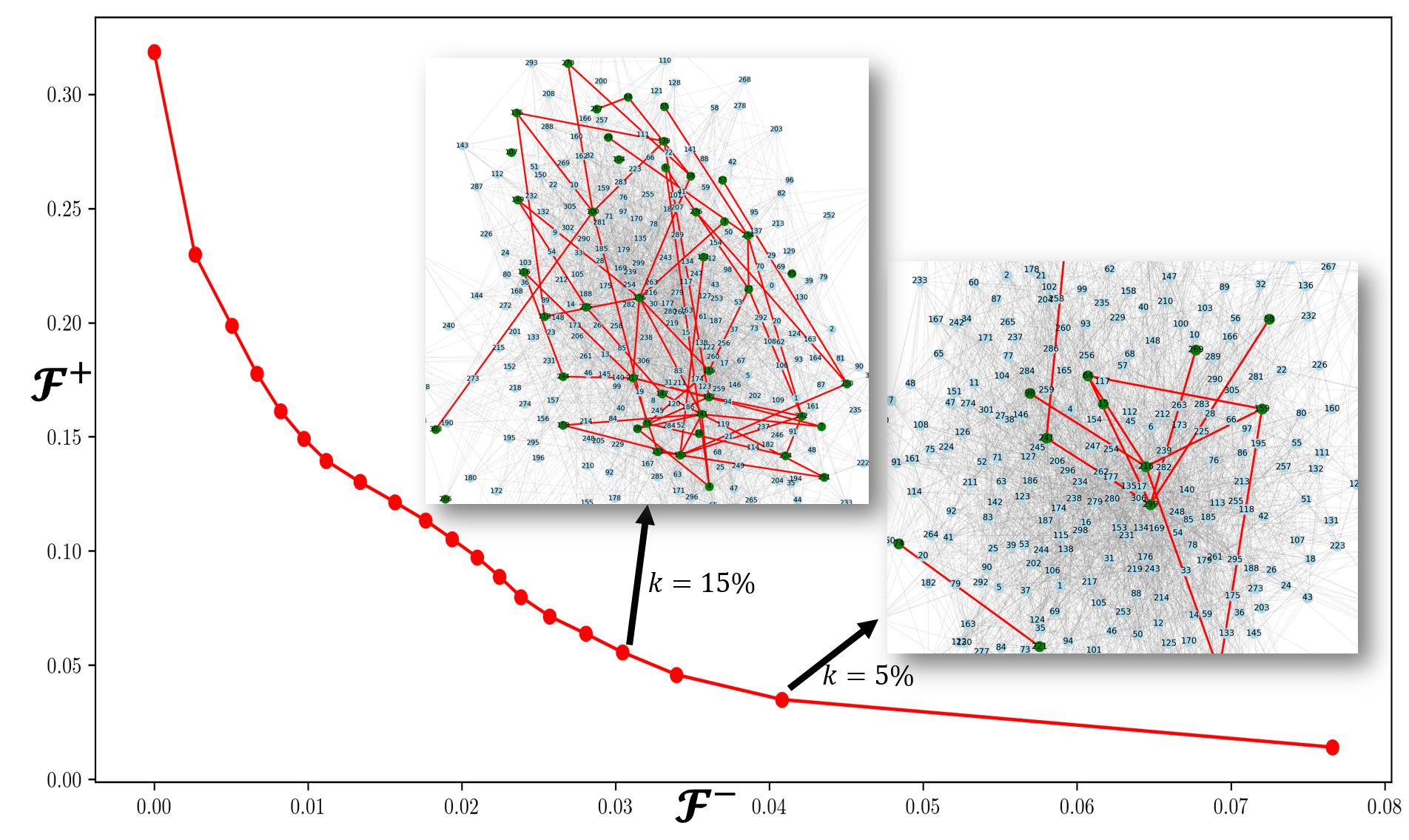}}
\caption{Positive over the negative fidelity based on the two selected subgraphs regarding the k\textcolor{red}{.}}
\label{fig_vis}
\end{figure}

figure \ref{fig_vis} illustrates the interpretability of the learned model by presenting extracted subgraphs at various density thresholds ($k$). These subgraphs, from the Pems04 dataset, exhibit consistent spatio-temporal patterns across the selected subgraph nodes regarding different values of $k$. This consistency underscores GINTRIP's ability to extract robust, time-aware subgraphs using the IB technique and learned prototypes. Our model detects soundly enriched subgraphs such that, by increasing their density, e.g. $k$, the inverse relationship between these two metrics can be observed.

\begin{figure}[htbp]
\centerline{\includegraphics[width=\linewidth,keepaspectratio]{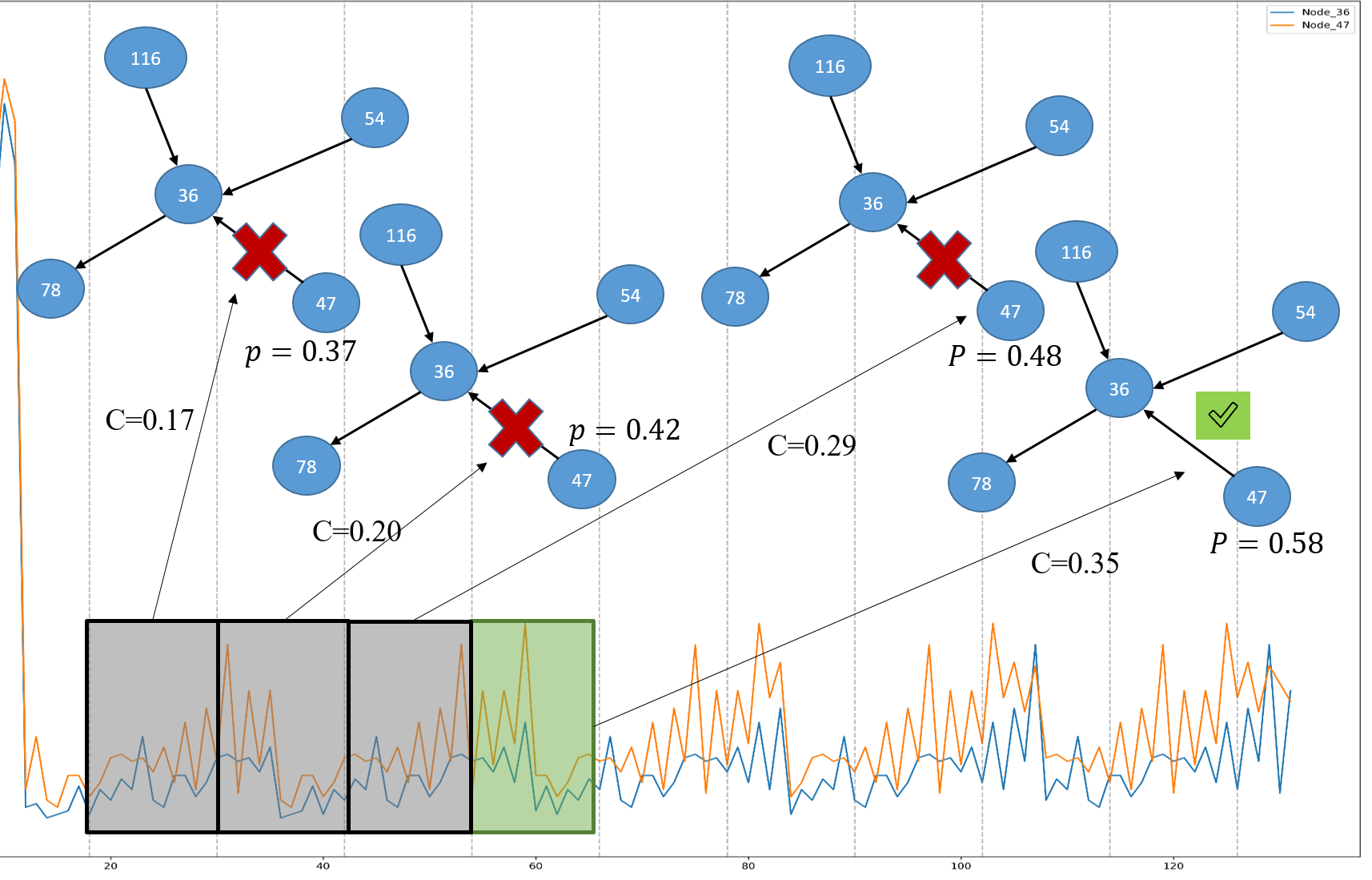}}
\caption{Explanation of the learned prototypes for a spatio-temporal prediction example. \textcolor{black}{The y-axis denotes traffic speed.}}
\label{fig_correlation}
\end{figure}

As a significant achievement, as opposed to STExplainer\cite{10.1145/3583780.3614871}, which
finds disjointed spatial and temporal subgraphs causes hampering the interpretation of the temporal interpretability, we introduce a unified subgraph that jointly
encoded spatial and temporal interpretability. To bolster this concept, we plotted learned subgraphs by network unfolded in time in figure~\ref{fig_correlation} for Pems04 dataset.
For the first three consecutive prediction windows, our
model did not select the edge $47 \rightarrow 36$, while in the
input graph, they are adjacent. The underlying reason
would be the temporal cross-correlation of the corresponding time series is not expressive enough. The
selection probabilities determined by the model truly
are proportional to the pairwise correlation.

\begin{figure}[!t]
    \centering
    \subfloat[Chicago crime rate prediction benchmark.]{%
        \includegraphics[width=0.45\textwidth]{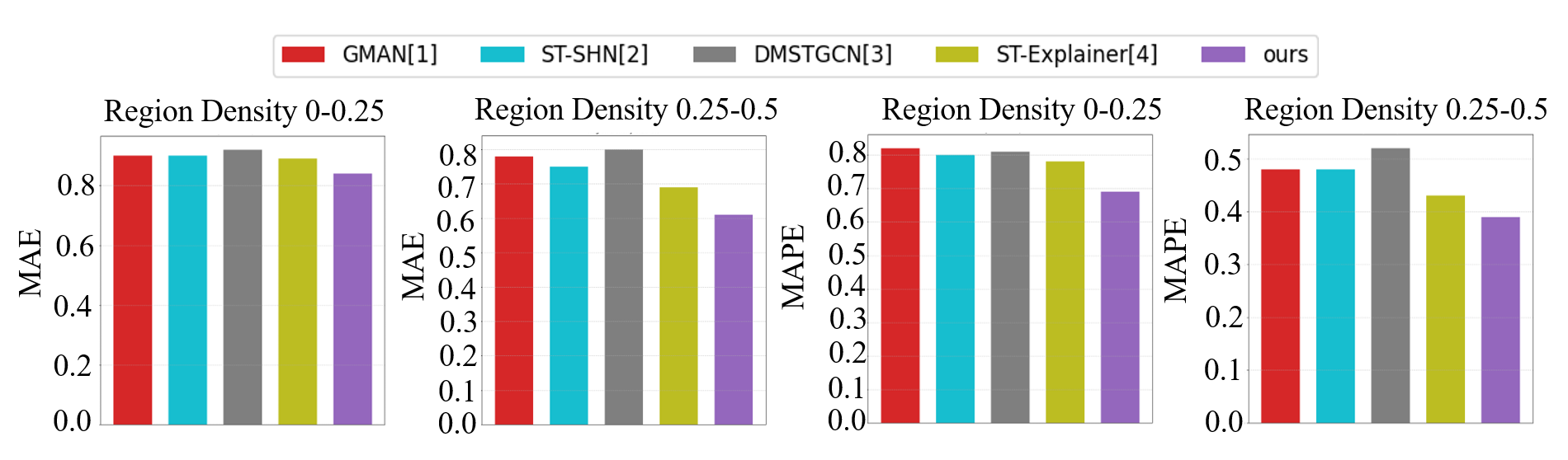}%
        \label{fig:sub_chicago}}
    \hfill
    \subfloat[New York City crime rate prediction benchmark.]{%
        \includegraphics[width=0.45\textwidth]{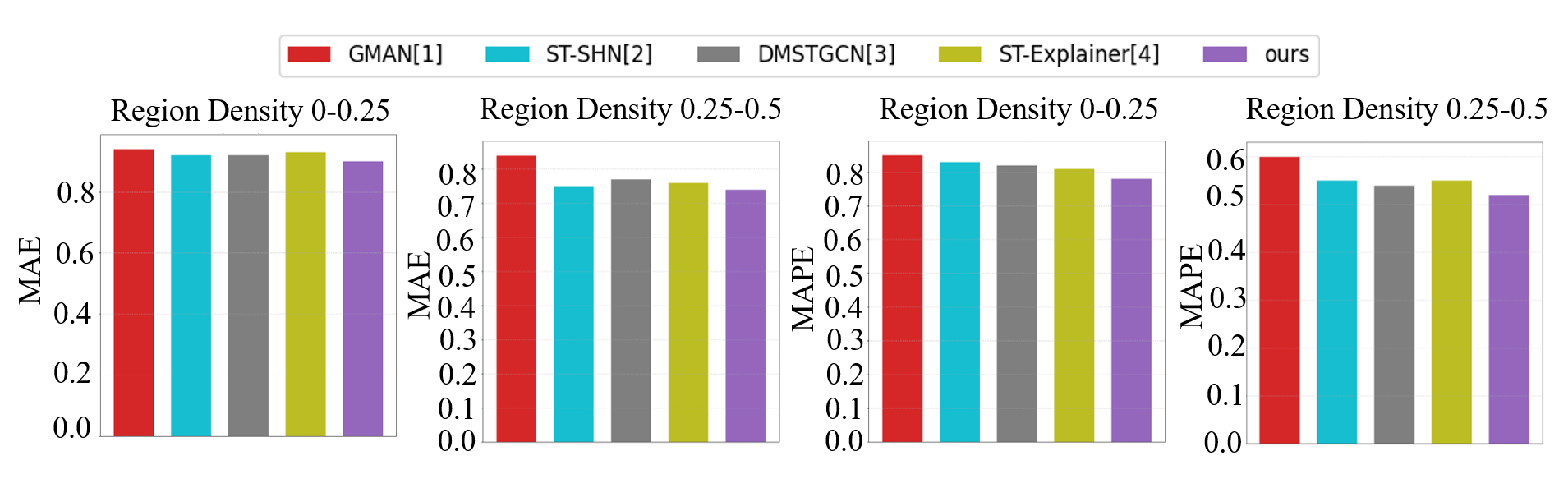}%
        \label{fig:sub_nyc}}
    \caption{Crime rate prediction benchmarks for Chicago and New York City.}
    \label{fig:crime_rate_benchmark}
\end{figure}

\subsection{Implementation Details}
\label{sec:implementation_details}
\textcolor{black}{
All experiments were implemented in PyTorch. To ensure reproducibility, a fixed random seed of $42$ was used for Python, NumPy, and PyTorch (including CUDA operations). Results are averaged through 10 parallel runs.
Model parameters were optimized using the Adam optimizer ~\cite{kingma2017adammethodstochasticoptimization} with default momentum parameters. The initial learning rate was scheduled using a cosine annealing policy. The exact learning rate and weight decay values are set to $0.001$ and $0.0001$, respectively. 
We referred to the $\beta$ hyperparameter as $\lambda_2$.The loss weighting hyperparameters, $\lambda$s, are $\lambda_1=1.0$, $\lambda_2=0.0005$, $\lambda_3=1.24291$, $\lambda_4=0.00007$, $\lambda_5= 0.000097$, respectively, their computation is mentioned in Section ~\ref{refcoeffadjusment}.
Prototypes were initialized from the Xavier method ~\cite{glorot2010understanding} and were updated jointly with the network parameters throughout training.
Models were trained for a maximum of $200$ epochs. Early stopping was applied based on the validation loss, with a patience of $20$ epochs, and the model corresponding to the best validation performance was retained for evaluation.
All experiments were conducted using two NVIDIA A100 GPUs. Pre-processing details are released in the Github repo.
}

\section{Ablation Study}

\subsubsection{Dense Area Analysis}
The ablation study has been conducted for the prediction in the dense area for both the NYC and Chicago datasets, which are depicted in figure \ref{fig:crime_rate_benchmark}. Consistently, in terms of prediction metrics, our model outperforms STExplainer\cite{Xia_2021}.

\subsubsection{Hyperparameter Analysis}\label{HPO}

\begin{figure}[htbp]
\centerline{\includegraphics[width=\linewidth,keepaspectratio]{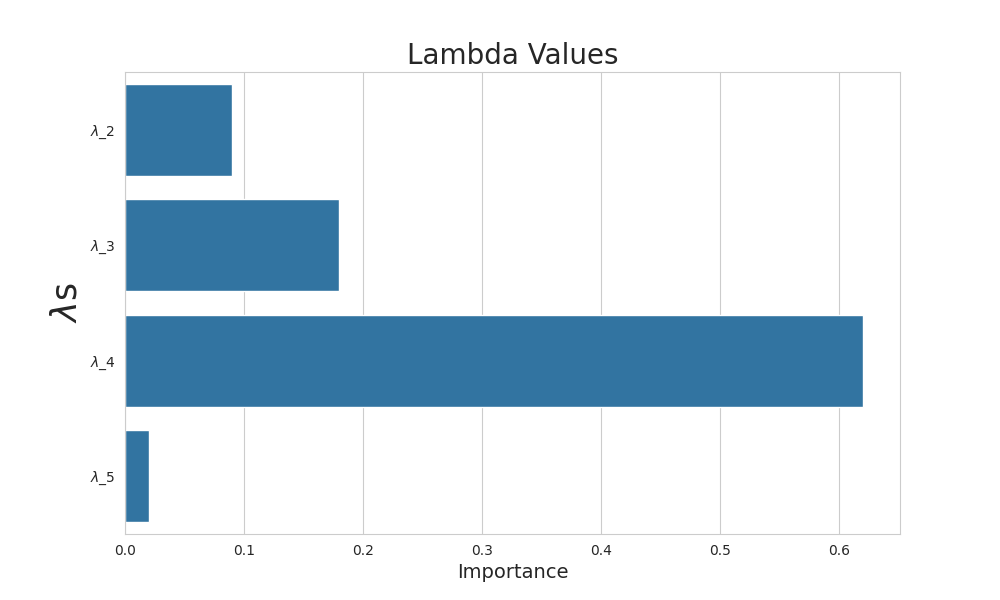}}
\caption{Hyperparameter importance ablation study}
\label{fig_hyper}
\end{figure}

In figure \ref{fig_hyper}, we present a comprehensive analysis examining the relative importance of various hyperparameters (denoted as $\lambda$s) and their impact on the final loss function described in equation \ref{final_loss}. Our findings demonstrate that GINTRIP consistently prioritizes the identification and extraction of subgraphs that exhibit three key characteristics: connectivity, representativeness, and interpretability. This is particularly evident through the hyperparameter $\lambda_4$, which plays a crucial role in this aspect of the model's behavior. Furthermore, we observed that $\lambda_3$ demonstrates notably high importance in our analysis, which can be directly attributed to its role in the learning process of interpretable prototypes. These prototypes are designed to closely mirror the structure and properties of the underlying subgraphs they represent. It's worth noting that since our primary objective focuses on regression tasks, we maintained the regression coefficient $\lambda_1$ at a constant value of 1 throughout our experiments, serving as a baseline for the other hyperparameters' relative contributions to the overall model performance. \textcolor{black}{To tune all hyperparameters, we adopted the approach proposed in \cite{groenendijk2020multilossweightingcoefficientvariations} (see Section~\ref{refcoeffadjusment}).}

\textcolor{magenta}{
\subsubsection{Prototype Module Importance}
\begin{table}[h]
    \centering
    \begin{tabular}{|c|c|c|c|c|}
        \hline
        & \multicolumn{2}{|c|}{PeMS04} & \multicolumn{2}{c|}{PeMS07} \\ \hline
        Run & MAE$\downarrow$ & RMSE$\downarrow$ & MAE$\downarrow$ & RMSE$\downarrow$ \\ \hline
        W/O Prototype & 18.85 & 29.26 & 20.28 & 33.71 \\ \hline
        \textbf{Ours} & \textbf{18.62} & \textbf{29.02} & \textbf{20.12} & \textbf{33.53} \\ \hline
    \end{tabular}
    \caption{Ablation study for the prototype module importance}
    \label{tab:ab2}
\end{table}
}
\textcolor{black}{Table \ref{tab:ab2} presents the ablation study conducted to assess the contribution of the proposed prototype module. The first row reports the performance when the module is disabled, resulting in higher errors across both datasets. In contrast, incorporating the module consistently improves MAE and RMSE on PeMS04 and PeMS07, demonstrating its effectiveness. These results highlight the necessity of the prototype module in improving predictive accuracy and validate its role as a key component of our framework.}

\subsubsection{Pseudo-label Robustness}


\begin{table}[h]
\small
\setlength{\tabcolsep}{4pt}  
\centering
\begin{tabular}{|l|c|c|}
\hline
\textbf{Method} & \textbf{Parameter} & \textbf{MAE} \\
\hline
Quantile & 0.80 & \underline{18.66}   \\ \hline
K-Means~\cite{kmeans} & 4 & 18.912  \\ \hline
DB-SCAN~\cite{K2024} & - & 18.928 \\ \hline
K-Means~\cite{kmeans} & 3 & 18.915  \\ \hline
Quantile & 0.70 & 18.75   \\ \hline
\textbf{Quantile (ours)} & \textbf{0.90} & \textbf{18.62}   \\ 
\hline
\end{tabular}
\caption{Sensitivity analysis of GINTRIP regarding quantile threshold on Pems04 dataset.}
\label{tab_robust}
\end{table}

\begin{figure}[htbp]
\centerline{\includegraphics[width=\linewidth,keepaspectratio]{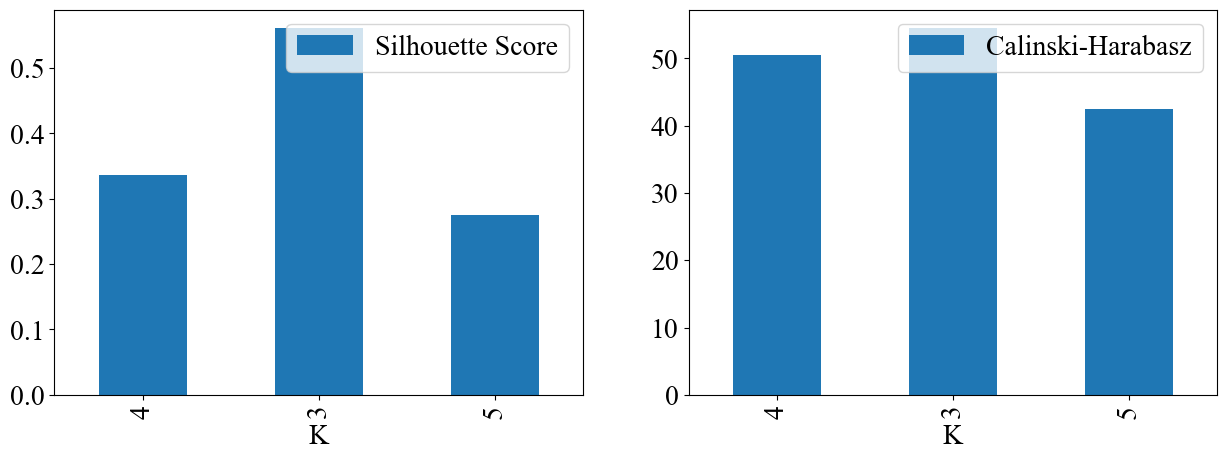}}
\caption{K-Means metric ablation study on Pems04 dataset.}
\label{fig_hyper_clus}
\end{figure}

\textcolor{black}{
As described in Section~\ref{problem}, we use a quantile-based function to determine pseudo-labels. In Table~\ref{tab_robust}, we ablate different threshold values. In addition, inspired by~\cite{K2024}, we implement K-means and DBSCAN to estimate the threshold in an unsupervised manner; the resulting clusterings are shown in Figure~\ref{fig_hyper_clus}. Overall, the different thresholding strategies yield comparable performance.
}




\section{Conclusion}

 In this study, we address the need for explainability in
spatio-temporal graph neural networks by introducing a novel framework called GINTRIP. This method not only predicts future
spatio-temporal signals accurately but also provides interpretable time-aware subgraphs. Our framework incorporates the prototype learning approach which employs variational
approximation for tractability. Additionally, we introduce a unified
Temporal-GNN that provides intrinsically explainable and robust spatio-temporal representations. Through comprehensive experiments, we examine the hypothesis that self-explainability causes predictability and robustness. Our results surpass existing state-of-the-art baselines in both predictive accuracy and explainability. In future research, we plan to investigate interpretability in hypergraphs.
\subsection{Limitations and Future Work}
While this paper provides insights into spatio-temporal neural networks, it implicitly assumes a static connectivity structure, overlooking the dynamics of changing connections. Future research could enhance our framework by incorporating dynamic causal models, allowing the model to adapt to variations in causal relationships over different connection times.

\appendices
\section{\break Summary of Notations}
{\color{black} 
For clarity, Table~\ref{tab:notation} summarizes the key notations used throughout this paper.
}
\begin{table*}[htbp]
\caption{Summary of Notations}
\label{tab:notation}
\centering
\begin{tabular}{|c|c|}
\hline
\textbf{Symbol} & \textbf{Definition} \\
\hline
\multicolumn{2}{|c|}{\textit{Graph Structure}} \\
\hline
$G_{in}$ & Input temporal graph \\
$V_{G_{in}}$ & Set of nodes in the input graph, $V_{G_{in}} = \{V_1, V_2, \ldots, V_N\}$ \\
$E_{G_{in}}$ & Set of edges in the input graph \\
$A_{G_{in}} \in \mathbb{R}^{N \times N}$ & Adjacency matrix of the input graph \\
$N$ & Total number of nodes in the graph \\
$G_{sub}$ & Extracted subgraph from $G_{in}$ via the IB principle \\
$G_p$ & Set of learnable prototypes, $G_p = \{\mathbf{z}_{G_p}^1, \mathbf{z}_{G_p}^2, \ldots, \mathbf{z}_{G_p}^M\}$ \\
\hline
\multicolumn{2}{|c|}{\textit{Temporal and Feature Variables}} \\
\hline
$X_t \in \mathbb{R}^{N \times D}$ & Dynamic node feature matrix at timestamp $t$ \\
$D$ & Dimensionality of node features \\
$T$ & Length of the observed time series history \\
$T'$ & Length of the forecast horizon \\
$W$ & Sliding window length ($W \leq T$) \\
$\boldsymbol{x} \in \mathbb{R}^W$ & Model input: sliding window $[y_{T-W+1}, \ldots, y_T]$ \\
$\boldsymbol{Y}$ & Prediction series (output) \\
$Y_{reg}$ & Ground-truth regression target \\
$\hat{Y}_{reg}$ & Predicted regression output \\
$Y_{cls}$ & Classification target (pseudo-labels) \\
\hline
\multicolumn{2}{|c|}{\textit{Latent Representations}} \\
\hline
$h_i$ & Latent representation of node $i$ from the GNN encoder \\
$z_i$ & Final representation of node $i$ in the extracted subgraph \\
$Z_{sub}$ & Pooled embedding of the subgraph $G_{sub}$ \\
$\mathbf{z}_{G_p}^m$ & The $m$-th learnable prototype vector \\
$Z_{G_p}$ & Concatenation of all prototype vectors $\{\mathbf{z}_{G_p}^1, \ldots, \mathbf{z}_{G_p}^M\}$ \\
\hline
\multicolumn{2}{|c|}{\textit{Prototype and Classification Variables}} \\
\hline
$M$ & Total number of prototypes ($M = K \times J$) \\
$K$ & Number of pseudo-classes (e.g., congested/non-congested) \\
$J$ & Number of prototypes allocated per class \\
$T_i$ & Threshold for classifying node $V_i$, defined as $Q_{0.1}(X_{:,i,:})$ \\
$S^m$ & Similarity score between $Z_{sub}$ and the $m$-th prototype $\mathbf{z}_{G_p}^m$ \\
\hline
\multicolumn{2}{|c|}{\textit{Subgraph Extraction Variables}} \\
\hline
$p_i$ & Probability of node $i$ being included in the subgraph $G_{sub}$ \\
$\lambda_i^{node}$ & Bernoulli sample $\lambda_i^{node} \sim \text{Bernoulli}(p_i)$ for node selection \\
$\epsilon$ & Gaussian noise, $\epsilon \sim \mathcal{N}(\mu_{h_i}, \sigma_{h_i}^2)$ \\
$\mu_{h_i}$ & Mean of the latent representation $h_i$ \\
$\sigma_{h_i}$ & Standard deviation of the latent representation $h_i$ \\
$P \in \mathbb{R}^{|V_G| \times 2}$ & Node probability assignment matrix for connectivity loss \\
$k$ & Subgraph density: ratio of selected nodes to total nodes \\
\hline
\multicolumn{2}{|c|}{\textit{Information-Theoretic Terms}} \\
\hline
$I(X; Y)$ & Mutual information between random variables $X$ and $Y$ \\
$H(X)$ & Entropy of random variable $X$ \\
$\beta$ & Lagrange multiplier controlling compression-prediction trade-off in IB \\
\hline
\multicolumn{2}{|c|}{\textit{Functions and Operators}} \\
\hline
$\gamma(\cdot, \cdot)$ & Similarity function (e.g., dot product) between $Z_{sub}$ and $G_p$ \\
$q_\phi(\cdot)$ & Variational approximation to posterior $p(Y_{cls} | \gamma(G_{sub}, G_p))$ \\
$q_\theta(\cdot)$ & Single-layer perceptron estimating $Z_{G_p}$ from $Z_{sub}$ \\
$f(\cdot)$ & Trained spatio-temporal predictive function \\
$\|\cdot\|_F$ & Frobenius norm \\
$\widehat{(\cdot)}$ & Row normalization operator \\
\hline
\multicolumn{2}{|c|}{\textit{Loss Functions and Weights}} \\
\hline
$\mathcal{L}$ & Total loss function \\
$\mathcal{L}_{reg}$ & Regression loss (MSE between $\hat{Y}_{reg}$ and $Y_{reg}$) \\
$\mathcal{L}_{cls}$ & Classification loss (cross-entropy for $Y_{cls}$) \\
$\mathcal{L}_{sub}$ & Upper bound on $I(G_{in}; G_{sub})$ for subgraph compression \\
$\mathcal{L}_{var}$ & Variational loss: $\text{MSE}(q_\theta(Z_{sub}), Z_{G_p})$ for prototype learning \\
$\mathcal{L}_{con}$ & Connectivity loss ensuring subgraph locality (Eq. \ref{final_representation}) \\
$\lambda_1, \ldots, \lambda_5$ & Weighting coefficients for each loss component \\
$\lambda, \lambda'$ & Constant coefficients in MI lower bounds (Eqs. \ref{eq:PGIB_final}, \ref{eq:UPB_CLS}) \\
\hline
\multicolumn{2}{|c|}{\textit{Evaluation Metrics}} \\
\hline
$\text{Fidelity}^{-}$ & Prediction change when using only the selected subgraph $G_i^{-}$ \\
$\text{Fidelity}^{+}$ & Prediction change when using the complement of subgraph $G_i^{+}$ \\
\hline
\end{tabular}
\end{table*}

\section*{Acknowledgment}
This work is funded by Innoviris within the research project TORRES under grant number 2022-RDIR-59c. AR and LDC acknowledge support from the Belgian Science Policy Office (BELSPO) under contract number B2/233/P2/PRECIP-PREDICT and through the FED-tWIN program (Prf-2020-017).

\bibliographystyle{ieeetr}
\bibliography{refrences.bib}

\begin{IEEEbiography}[{\includegraphics[width=1in,height=1.25in,clip,keepaspectratio]{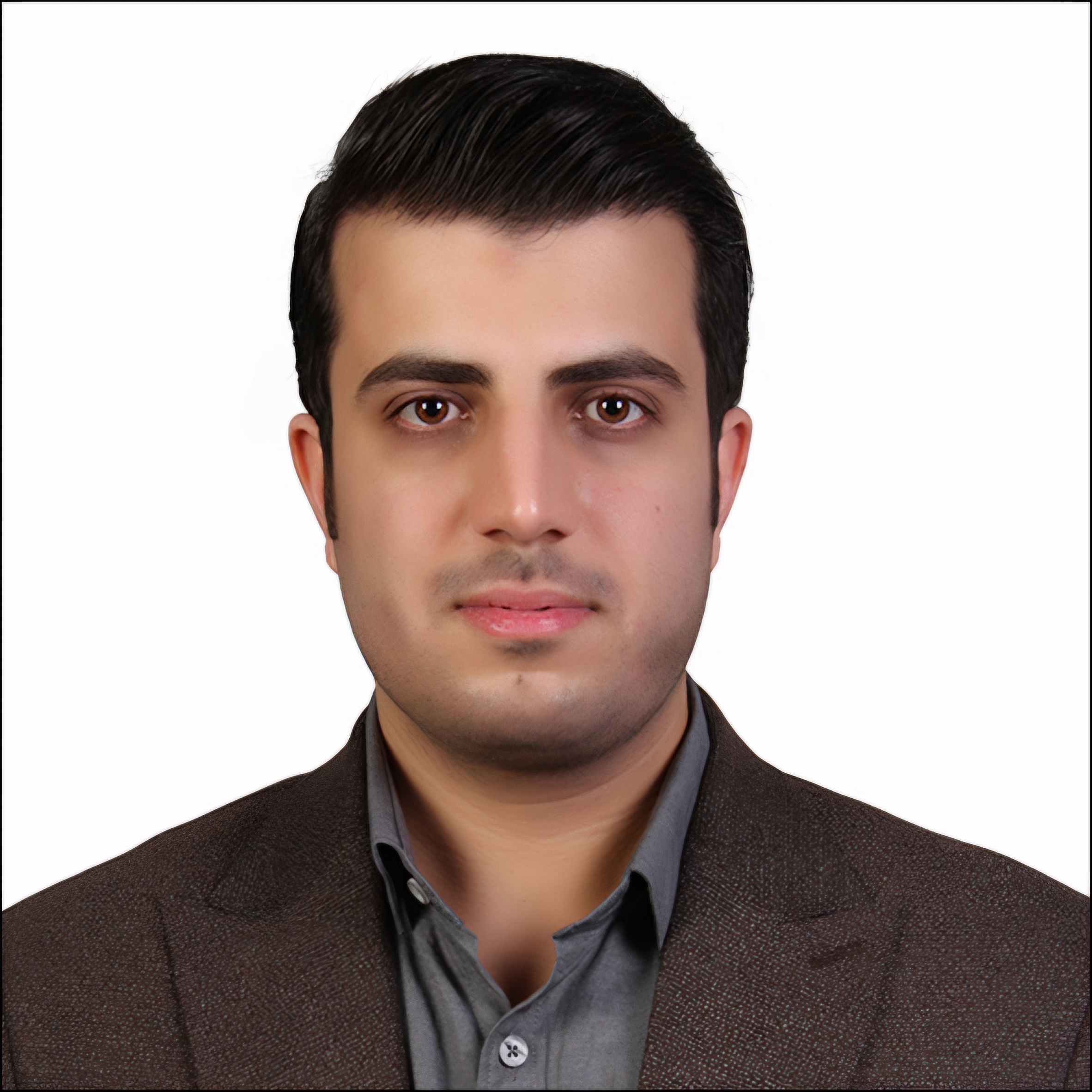}}]{Ali Royat} 
received his M.Sc. degree in Electrical Engineering at the Sharif University of Technology, Tehran, Iran, in 2020. His Background lies in applying
deep learning methods to Computer Vision and signal processing. He is currently a \textcolor{black}{{researcher}} in the Department of
Electronics and Informatics, (VUB), Brussels, Belgium.
\end{IEEEbiography}

\begin{IEEEbiography}
[{\includegraphics[width=1in,height=1.25in,clip,keepaspectratio]{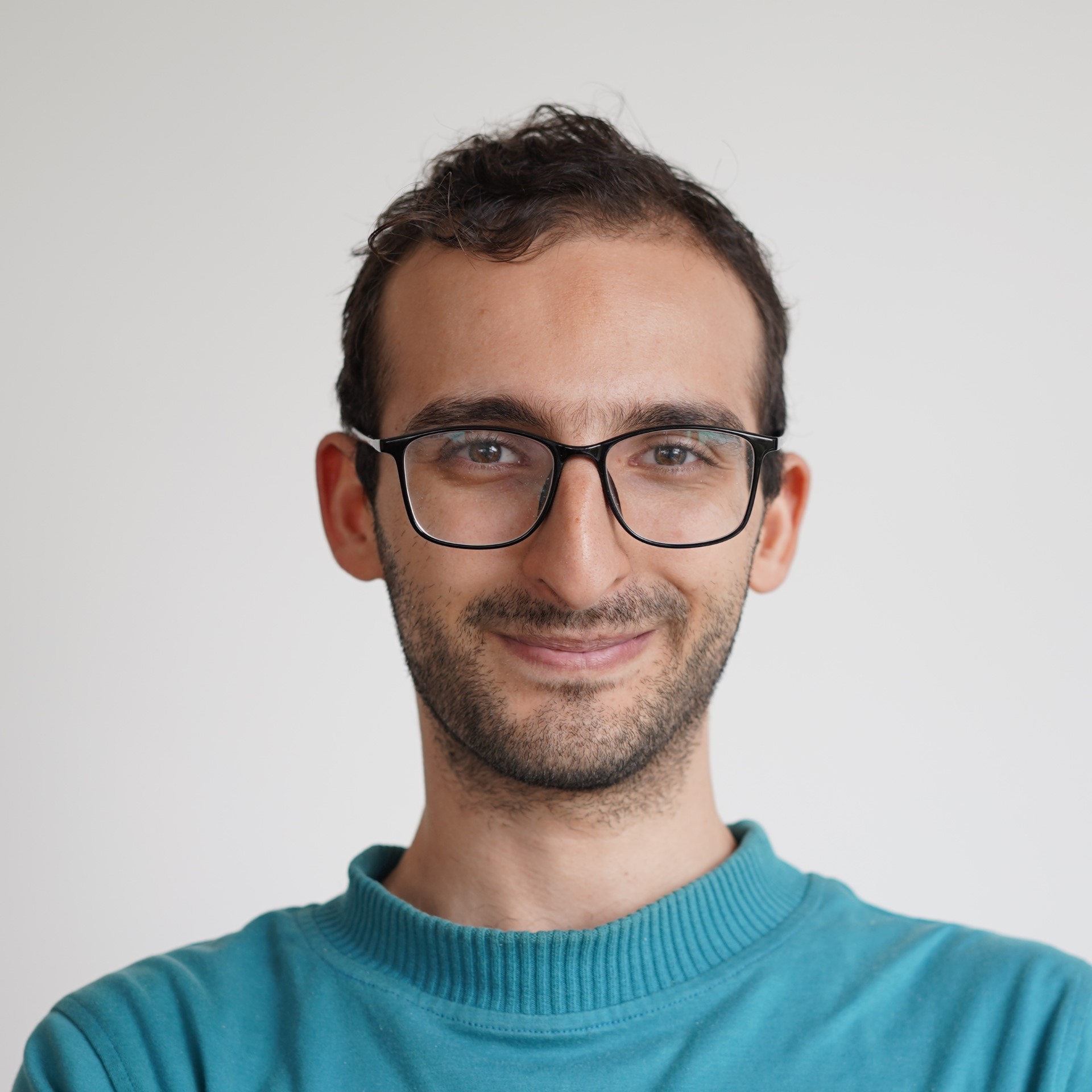}}]{Seyed Mohamad Moghadas} received the M.Sc. degree of Computer Science from Amirkabir University of Technology Tehran, Iran, in 2023. He was a Software Engineer at Mahsan Company. He is currently a PhD student in the Department of
Electronics and Informatics, (VUB), and IMEC,
Brussels, Belgium. He investigated the use of various graph neural networks in the transportation domain. His research interests
include spatio-temporal data analysis and sparse representations in time-series analysis.
\end{IEEEbiography}

\begin{IEEEbiography}[{\includegraphics[width=1in,height=1.25in,clip,keepaspectratio]{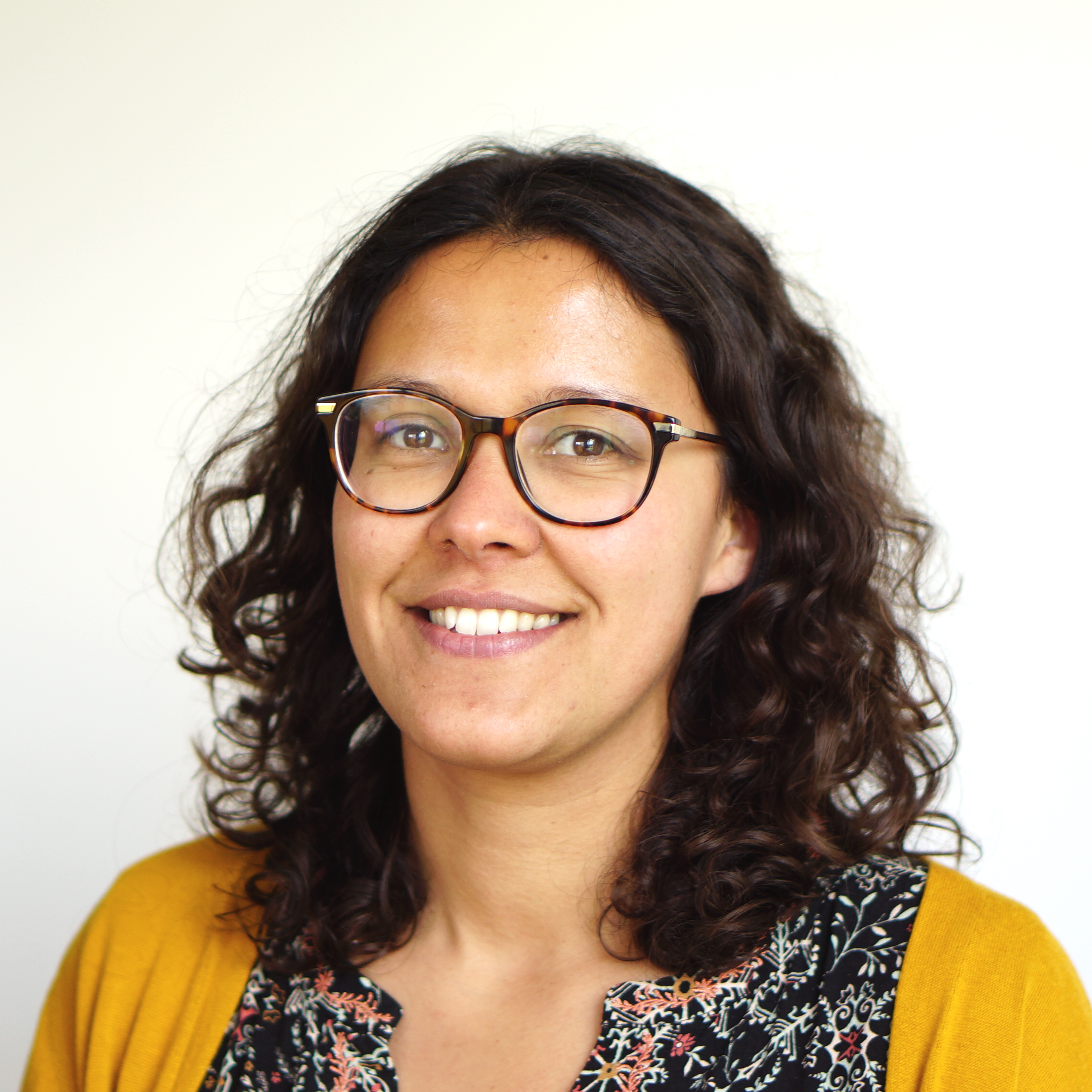}}]{Lesley De Cruz} is an assistant professor at the Department of Electronics and Informatics (ETRO) at Vrije Universiteit Brussel (VUB) and a senior researcher at the Royal Meteorological Institute of (RMI) of Belgium.
She obtained a PhD in Physics at Ghent University in 2011 and joined the RMI in 2012. As of 2021 she holds the joint “FED-tWIN” professorship between RMI and VUB, funded by BELSPO (Belgian Science Policy Office). She coordinates the RMI’s Seamless Prediction Programme, which aims to deliver accurate weather forecasts by leveraging high-resolution multisensory observations, innovative nowcasting, probabilistic modelling and artificial intelligence (AI).  \end{IEEEbiography}

\begin{IEEEbiography}[{\includegraphics[width=1.2in,height=1in,clip,angle=-90]{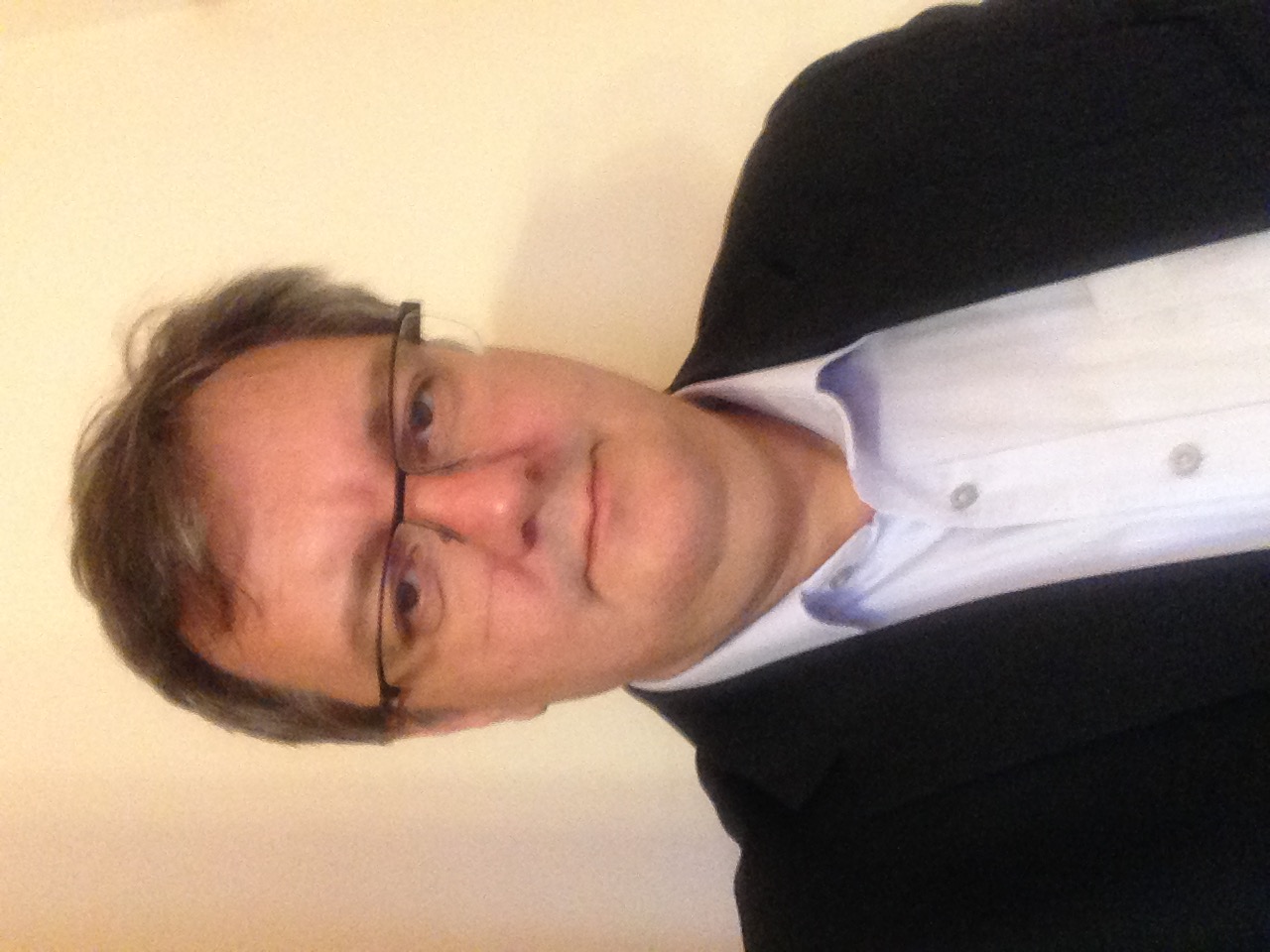}}] {{Adrian Munteanu}}
(M’07) is professor at the Electronics and Informatics (ETRO) department of the Vrije Universiteit Brussel (VUB), Belgium. He received the MSc degree in Electronics and Telecommunications from Politehnica University of Bucharest, Romania, in 1994, the MSc degree in Biomedical Engineering from University of Patras, Greece, in 1996, and the Doctorate degree in Applied Sciences (Maxima Cum Laudae) from Vrije Universiteit Brussel, Belgium, in 2003. In the period 2004-2010 he was post-doctoral fellow with the Fund for Scientific Research – Flanders (FWO), Belgium, and since 2007, he is professor at VUB.
His research interests include image, video and 3D graphics compression, deep learning, distributed visual processing, error-resilient coding, and multimedia transmission over networks.
He is the author of more than 400 journal and conference publications, book chapters, and contributions to standards, and holds 7 patents in image and video coding. He is the recipient of the 2004 BARCO-FWO prize for his PhD work, the (co-)recipient of the Most Cited Paper Award from Elsevier for 2007, and of 10 other scientific prizes and awards. He served as Associate Editor for IEEE Transactions on Multimedia and currently serves as Associate Editor for IEEE Transactions on Image Processing.
\end{IEEEbiography}

\newpage


\EOD

\end{document}